\documentclass[conference]{IEEEtran}
\IEEEoverridecommandlockouts
\usepackage[numbers]{natbib}
\usepackage{amsmath,amssymb,amsfonts}
\usepackage{graphicx}
\usepackage{textcomp}
\usepackage{xcolor}
\usepackage{algpseudocode}
\usepackage{algorithm}
\usepackage{float}
\usepackage{booktabs}
\usepackage[left=0.68in,right=0.68in,top=0.7in]{geometry}

\def\BibTeX{{\rm B\kern-.05em{\sc i\kern-.025em b}\kern-.08em
    T\kern-.1667em\lower.7ex\hbox{E}\kern-.125emX}}

\begin{document}

\title{FedLE: Federated Learning Client Selection with Lifespan Extension for Edge IoT Networks}

\author{\IEEEauthorblockN{Jiajun Wu}
\IEEEauthorblockA{\textit{Electrical and Software Engineering} \\
\textit{University of Calgary}\\
Calgary, Canada \\
jiajun.wu1@ucalgary.ca}
\and
\IEEEauthorblockN{Steve Drew}
\IEEEauthorblockA{\textit{Electrical and Software Engineering} \\
\textit{University of Calgary}\\
Calgary, Canada \\
steve.drew@ucalgary.ca}
\and
\IEEEauthorblockN{Jiayu Zhou}
\IEEEauthorblockA{\textit{Computer Science and Engineering} \\
\textit{Michigan State University}\\
East Lansing, USA \\
jiayuz@msu.edu}

}

\maketitle

\begin{abstract}
Federated learning (FL) is a distributed and privacy-preserving learning framework for predictive modeling with massive data generated at the edge by Internet of Things (IoT) devices. 
One major challenge preventing the wide adoption of FL in IoT is the pervasive power supply constraints of IoT devices due to the intensive energy consumption of battery-powered clients for local training and model updates. 
Low battery levels of clients eventually lead to their early dropouts from edge networks, loss of training data jeopardizing the performance of FL, and their availability to perform other designated tasks. 
In this paper, we propose FedLE, an energy-efficient client selection framework that enables lifespan extension of edge IoT networks. 
In FedLE, the clients first run for a minimum epoch to generate their local model update. 
The models are partially uploaded to the server for calculating  similarities between each pair of clients. 
Clustering is performed against these client pairs to identify those with similar model distributions. 
In each round, low-powered clients have a lower probability of being selected, delaying the draining of their batteries.
Empirical studies show that FedLE outperforms baselines on benchmark datasets and lasts more training rounds than FedAvg with battery power constraints. 
\end{abstract}


\section{Introduction}

The pervasive deployment of Internet of Things (IoT) devices has resulted in large-scale personal data generated at the network edge \cite{wu2023topology}. 
The edge-based IoT data cannot be uploaded to a centralized server for machine learning due to privacy concerns and resource limitations. 
Federated learning (FL) allows distributed clients to collaboratively learn a model without uploading their local training data, providing an ideal solution for privacy-privacy distributed learning in edge computing.
In reality, challenges exist for effective FL, as IoT networks are largely battery-powered and heterogeneous, with diversified battery capacities and potentially short lifespans.
What makes the problem worse is that IoT devices consume a significant amount of energy during the training process of FL. A portion of devices would deplete their energy earlier than others and drop out of the network. 
These dropouts reduce the number of available clients, which in turn limits access to the training data.
With the vanilla FedAvg~\cite{mcmahan2017communication} performing client selection uniformly at random, the heterogeneous energy profile and status of clients are not considered. 
A critical client with low remaining energy has an equal probability of being selected compared to a non-critical client for a new round of data aggregation. 
These facts may lead to early drops of critical clients, overall lower network availability, and consequently, poor model performance and longer convergence time.

Due to the potential loss of critical clients, client selection becomes a major concern.
Existing studies primarily focused on optimizing energy efficiency with bandwidth allocation in wireless networks.
\citet{yu2022jointly} investigated the tradeoffs between maximizing the number of selected clients and minimizing total energy consumption, and selecting clients with an optimized allocation of CPU resources for FL local training and transmission power for model updates.
The impact on learning performance with varying temporal client selection patterns was discussed by \cite{xu2021client}. 
A stochastic optimization problem was then formulated for joint bandwidth allocation and client selection with a long performance guarantee. 
FedMCCS \cite{abdulrahman2021fedmccs} aimed to maximize the number of clients selected each round while considering client resource budgets and capabilities, such as CPU, memory, and energy. 
Although the client selection rate and test accuracy have been improved in certain scenarios, existing studies did not factor in an unreliable edge computing environment to maximize the lifespan of the entire edge networks with more available clients.

\begin{figure}[t]
    \centering
    \includegraphics[width=0.48\textwidth]{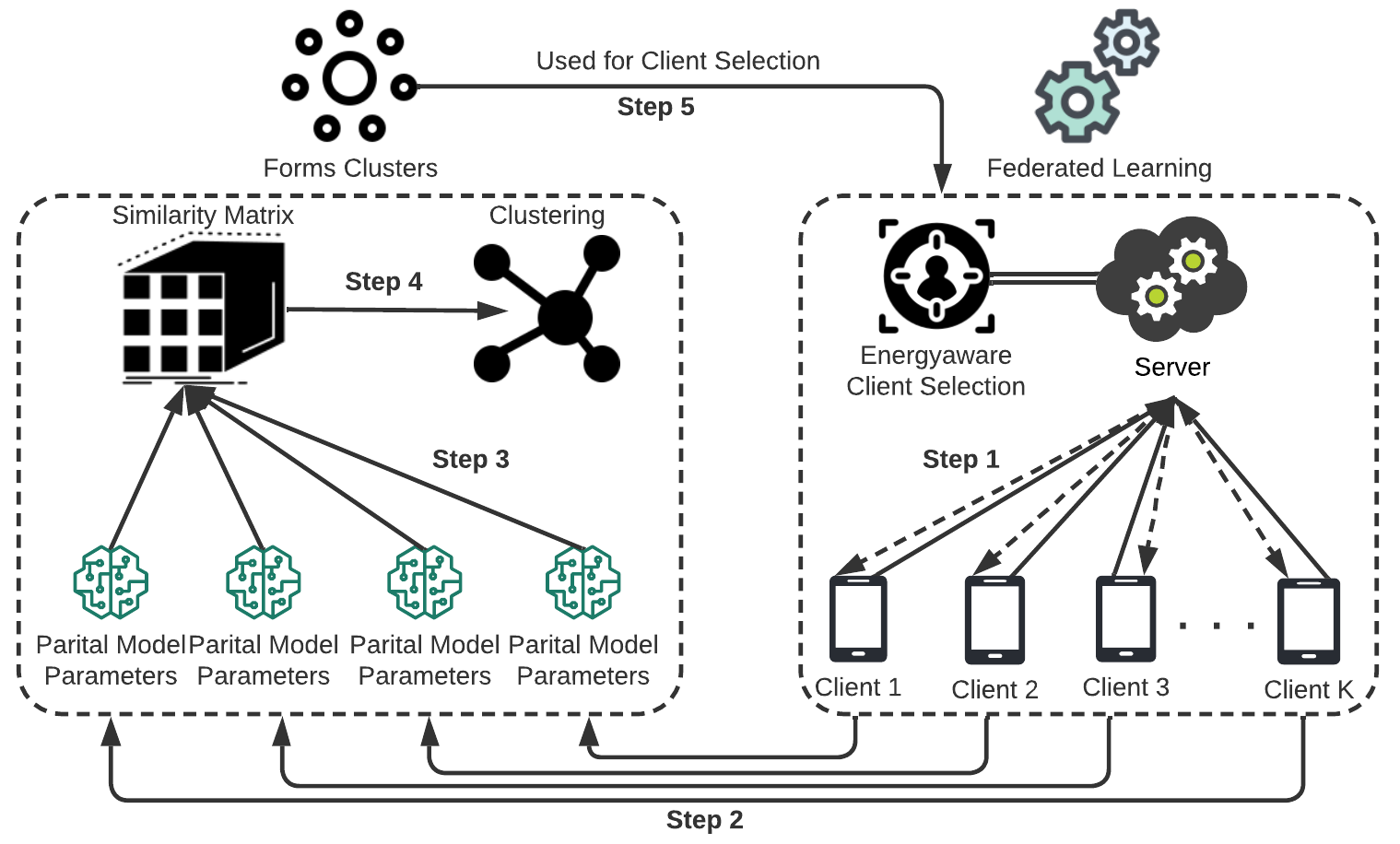}
    \vspace{-0.1in}
    \caption{Overview of the FedLE client selection framework. 1) Before the initial training round, all clients run for a minimum epoch and upload partial model weights to the server. 
2) Server computes the similarity matrices. 
3) Represent every client by comparing each with the lowest similarity pair in the weight space. 
4) Cluster the clients based on their weights. 
5) Dynamically set chances of selection based on the cluster size. }
    \label{fig:FedLE-overview}
    \vspace{-0.1in}
\end{figure}

In this paper, we propose FedLE, a clustering-based client selection method for FL that accelerates training for heterogeneous data.
The intuition of FedLE is from model similarity, where clients with similar data tend to provide similar models during FL. 
Therefore, a sampling strategy that prioritizes the selection of clients with very different models can be an energy-efficient approximation to cut down potentially redundant model uploads and thus accelerate the learning processing of FL. 
With that in mind, FedLE forms clusters based on clients' local model weights following the steps shown in Fig. \ref{fig:FedLE-overview}, where the server computes similarity matrices of client model weights and clusters the clients. The probability of a client being selected will depend on the size of its cluster. With this approach, we are not compromising client privacy, directly accessing client data, or acquiring new information from the clients. Compared to existing studies, FedLE has the following contributions:
\begin{itemize}
    \item FedLE balances the energy consumption and makes all clients last longer. To our knowledge, this is the first work considering edge network lifespan extension for FL.
    \item Our proposed method uses partial model layers to build the similarity matrix for clustering, which proves to be far more energy-efficient than using full models.
    \item We demonstrate through empirical studies that in most cases, the creation of similarity matrix and clustering only needs to be performed once, therefore only adding a small overhead while still outperforming baseline methods.
\end{itemize}

\section{Related Work}\label{sec:related-work}
FedAvg \cite{mcmahan2017communication} demonstrated that random client selection is effective when all clients are available and adequately trained. To improve the performance of FL with heterogeneous data, FedProx was proposed \cite{li2020federated}, introducing a proximal term to restrict the local updates to be closer to the global model.

\noindent\emph{Energy-efficient client selection} has become one of the popular topics in FL. The power-of-choice \cite{cho2020client} client selection framework prioritized communication and computation efficiency. \citet{nishio2019client} proposed a client selection protocol for FL called FedCS that accounts for both energy resources and heterogeneous data in mobile edge computing (MEC) environments. While existing methods have improved energy efficiency in certain scenarios, they did not consider extending the lifespan of edge and IoT networks to balance the energy consumption of the clients. In comparison, our proposed FedLE framework takes into account the total number of rounds for FL in an attempt to extend the effective lifespan of the entire edge network to retain the data sources and to help perform the primary tasks of the IoT devices. In FedSim \cite{palihawadana2022fedsim}, pairwise client similarity was used for clustering as well, but it made full use of the local model parameters of the client and took clustering as a post-FL processing technique. As part of our proposed method, we utilize only partial local model parameters in a one-time manner for client selection. \citet{wang2021edge} proposed an FL scheme that removed ineffective models using cosine similarity that clusters devices by vectorizing local and global model parameters. It also required the full model parameters and did not take energy into account. Our proposed method simply computes similarity using local model parameters and requires less computation and communication power.

\section{The FedLE Client Selection Framework}
In this section, we introduce the FedLE client section framework with the problem formulation.
\subsection{Constrained battery power}\label{sec:problem-form}
We first define an edge-based FL scenario with a set of battery-powered clients $\mathbb{C}$, where $\left| \mathbb{C} \right| = K$. All clients communicate with an aggregation server $S$. We use the communication rounds in FL to represent the current time in the system. Specifically, for the $k$th client $c_k \in \mathbb{C}$, define its battery level as the remaining battery power at Round $t$ for active use, denoted by $b_t^k \in [0, 1]$, where $0$ stands for completely drained battery, and $1$ for a fully charged battery. As long as a client is turned on, sitting idle will consume the device's battery. We define the standby battery discharge rate of $c_k$ as the amount of $b_t^k$ dropped in each epoch of FL training time, denoted by $r_t^k$. For the clients
selected for model updates, transmission to the server will result in additional consumption of the battery. We define the battery discharge caused by each model update transmission to be $s_t^k$. For any battery discharge resulting from other communications, we denote the amount as $a_t^k$ for $c_k$. The total level at Round $t + 1$ for $c_k$ will be:
\begin{align*}
    b_{t+1}^k = b_t^k - r_t^k - s_t^k - a_t^k.
\end{align*}
Edge devices typically have their primary tasks besides FL. Instead of allowing the FL process to completely drain the device battery, we set a critical battery level $\delta$. If the client battery level is lower than $\delta$, the FL process will halt to allow the edge IoT device to complete its primary tasks.


\subsection{Selecting statistically diverse clients}
\label{sec:similarity}
In FL, large numbers of clients can carry highly similar data in distribution. Typical examples include IoT sensors in industrial environments with close proximity, surveillance cameras on the same street, and smartphones of users with matching preferences. 
It is necessary to avoid selecting such clients in consecutive rounds during the FL process. When a majority of clients share data with similar distributions, the training results would likely lead to overfitting to the majority class. 
On the other hand, the energy constraints of the clients limit their total rounds of training: the rounds of the clients with limited remaining energy are numbered. Therefore, the opportunity to participate in training is valuable. It is important to select clients with diverse data distribution that boosts the convergence of the global model. We plan to compare the local model weights of each client to form a similarity matrix and cluster the clients based on their weight.

\algnewcommand\algorithmicforeach{\textbf{for each}}
\algdef{S}[FOR]{ForEach}[1]{\algorithmicforeach\ #1\ \algorithmicdo}

\begin{algorithm}[htb]
\caption{The Algorithm Building the Similarity Matrix }\label{alg:Matrix}
\begin{algorithmic}[1]
\If{initial round}
  \ForEach{client $c_k \in \mathbb{C}$}
    \State locally train client $c_k$ for 1 epoch
    \State $c_k$ sends model updates $w_k$ to the server
    \State Server flattens \textbf{partial} $w_k$
  \EndFor
  \State Server compares the dot products for each pair of the $K$ clients to build a $K \times K$ similarity matrix
  \State Server sums up the value for each row in the $K \times K$ matrix into a one-dimension list of size $K$
\EndIf
\end{algorithmic}
\end{algorithm}

\subsection{Efficiently building the similarity matrix}
The major challenges of building a similarity matrix in practice include the overhead introduced by comparing every two of the $K$ clients for their gradients. A total of $\binom{K}{2} / 2$ comparisons of full model updates are required for creating the similarity matrix. Although this process is completed on the server side, it introduces significant computational cost and latency. The problem worsens when $K$ and the model gradient matrix are large. To address these challenges in building the similarity matrix, we partially load model parameters to the server when updating the matrix. Based on our observations, there is a significant difference between the local client models in terms of model weights and bias, particularly in the first and last layers of a convolutional neural network (CNN) model. Therefore, instead of using the entire model to formulate the matrix, we only use the model parameters of the first convolution layer and the last fully connected layer. Suppose after trimming the model, for each gradient matrix of $c_k$, we have a $D \times D$ gradient matrix.

To prove the effectiveness of using partial models, we design an experiment to show the cosine similarity of each client using partial model parameters in the FL process. In FedAvg setting, we train a CNN model with MNIST \cite{lecun1998mnist} for a total of 30 rounds and CIFAR-10 \cite{krizhevsky2009learning} dataset for a total of 20 rounds, each with 15 clients. Every round, each client computes the cosine similarity between its local model weights and all other clients to get $k$ cosine similarity values. We then formulate a $K \times K$ matrix where each cell corresponds to the similarity scores of the two clients, calculated as below:
\begin{equation}
\label{eq:dot-product}
sim(c_i,c_j) = \mathbf{w_i} \cdot \mathbf{w_j} =\sum_{d_i=1}^{D}\sum_{d_j=1}^{D} w_i^{d_i} w_j^{d_j},
\end{equation}
where $\mathbf{w_i}$ and $\mathbf{w_j}$ are the gradient matrices of client $c_i$ and $c_j$, respectively. The variables $w_i^{d_i}$ and $w_j^{d_j}$ represent the cell values in the gradients of $\mathbf{w_i}$ and $\mathbf{w_j}$. Applying Equation (\ref{eq:dot-product}), we visualize the similarity matrix in the format of heatmaps for all the matrices in each round with full and partial weights. The results are shown in Fig. \ref{fig:sim-full-weight}.

\subsubsection{The importance of the initial similarity matrix}
\begin{figure}[t]
    \centering
    \includegraphics[width=0.49\linewidth]{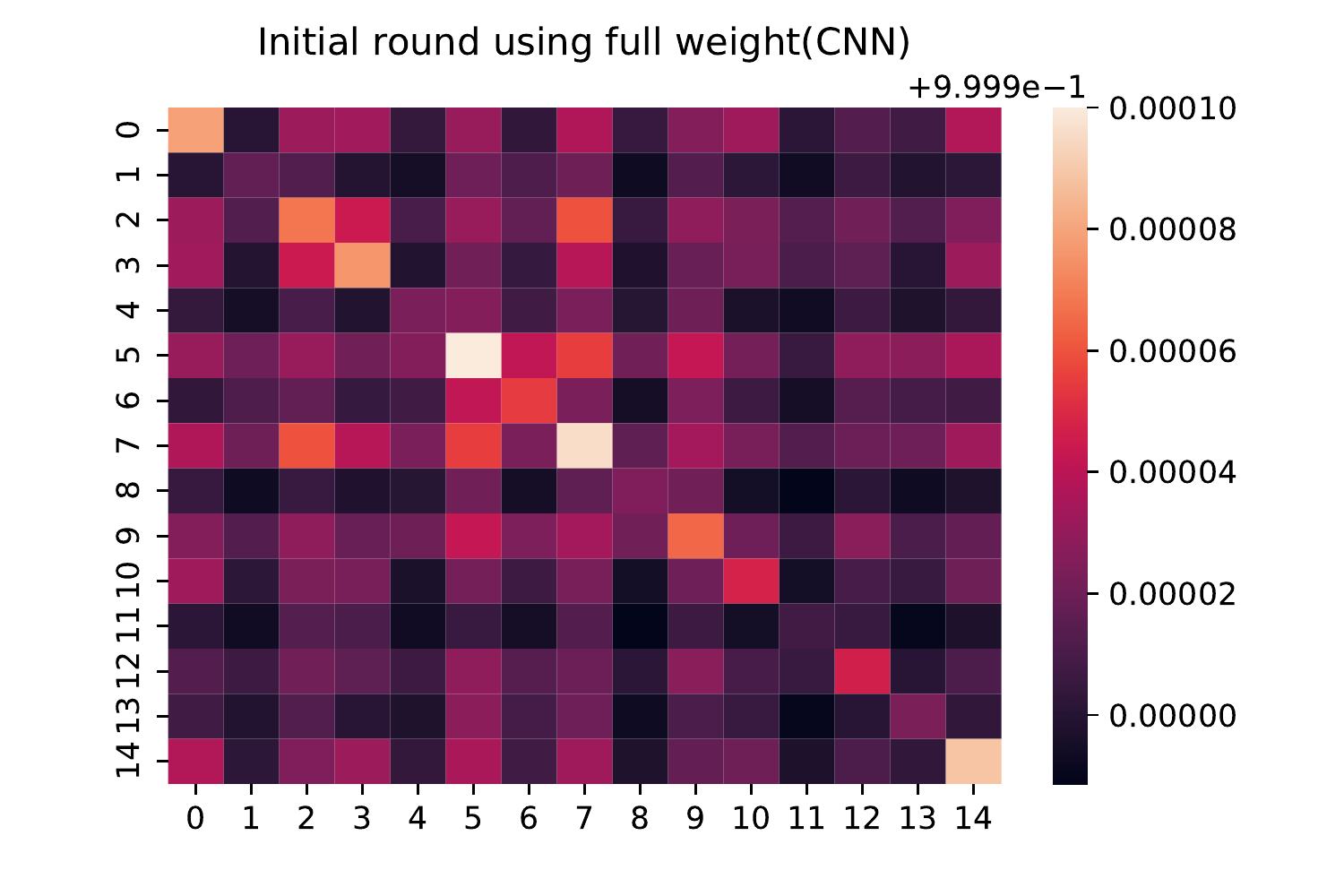}
    \includegraphics[width=0.49\linewidth]{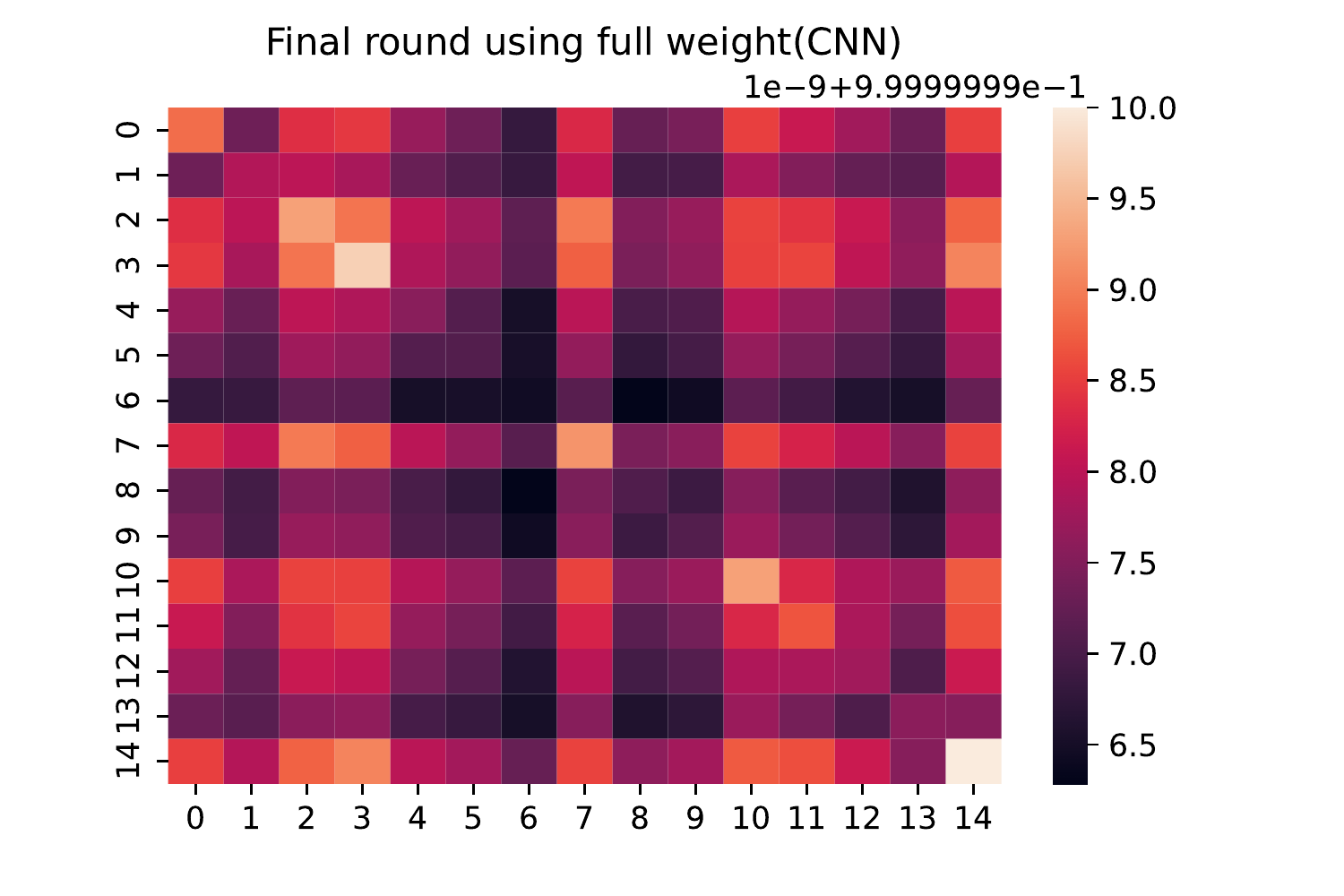}
    \includegraphics[width=0.49\linewidth]{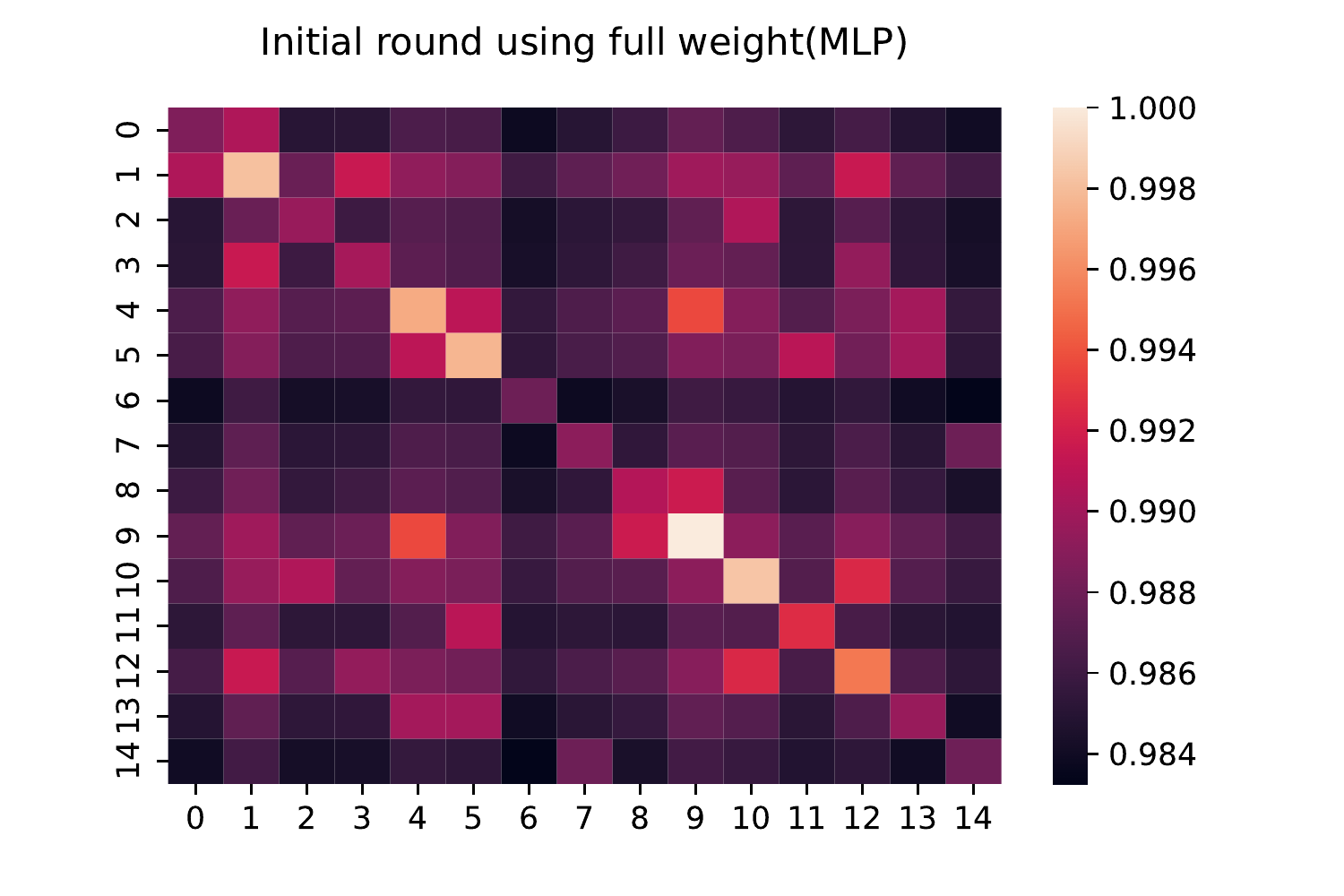}
    \includegraphics[width=0.49\linewidth]{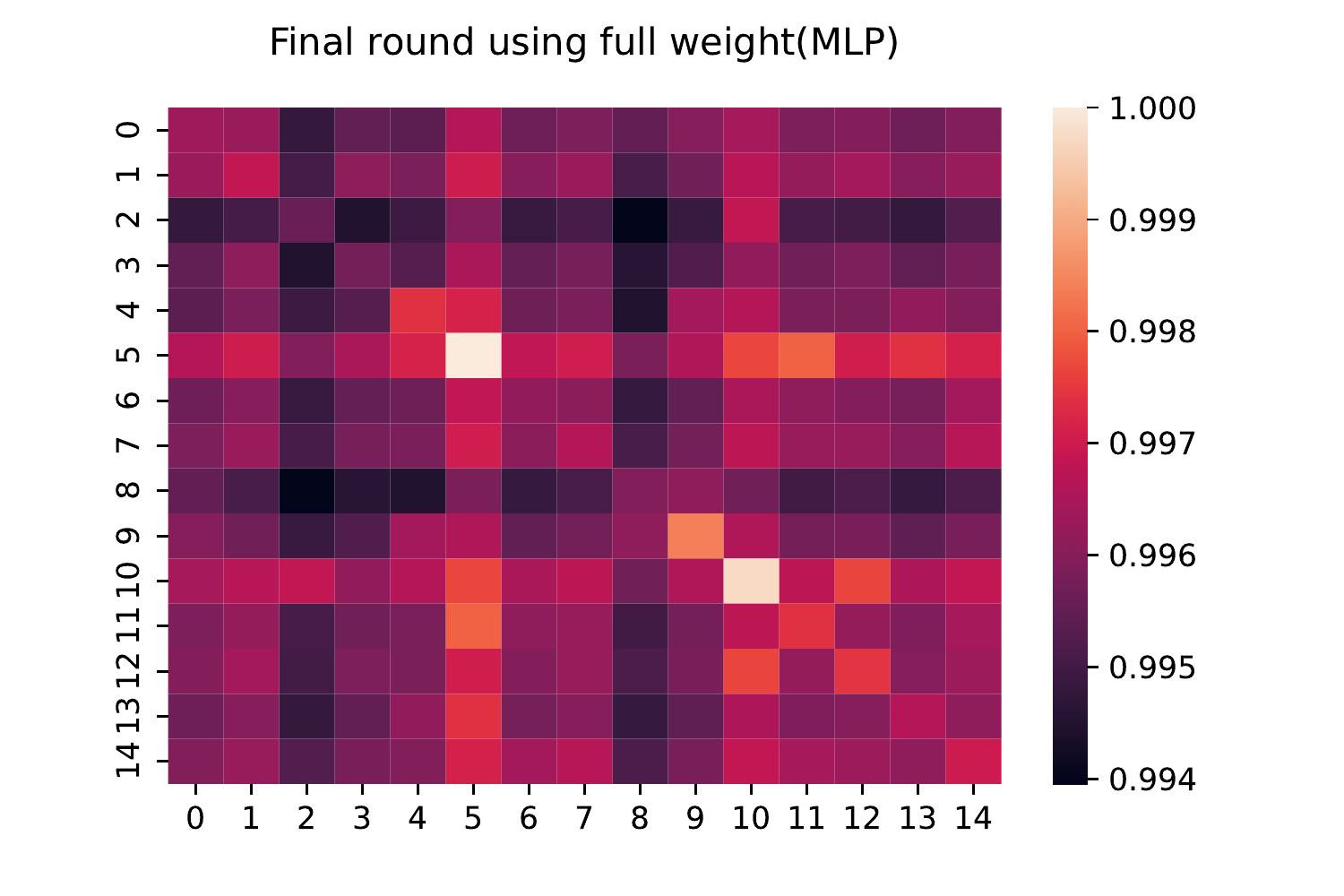}    
    \vspace{-0.2in}
    \caption{The similarity matrix using full model weights. (Top row: CNN model's initial round and final round matrix. Bottom row: MLP model's initial round and final round matrix.)} 
    \vspace{-0.2in}
    \label{fig:sim-full-weight}
\end{figure}

By observing the heatmaps from Fig. \ref{fig:sim-full-weight}, there is no significant change in the pattern. We also learn that as the number of rounds increases, the pattern of the similarity matrix remains relatively stable. There is no significant difference between the consecutive matrices and even latter matrices. We believe this is because the distribution of local data on each client is often stable and not likely to change dramatically. It is evident from this experiment that the use of the similarity matrix is valid and the matrix does not need to be updated frequently. 

\subsubsection{Robustness with different levels of non-IIDness and imbalanced data}
\begin{figure}[!tbp]
  \centering
  \begin{minipage}[b]{0.24\textwidth}
    \includegraphics[width=\textwidth]{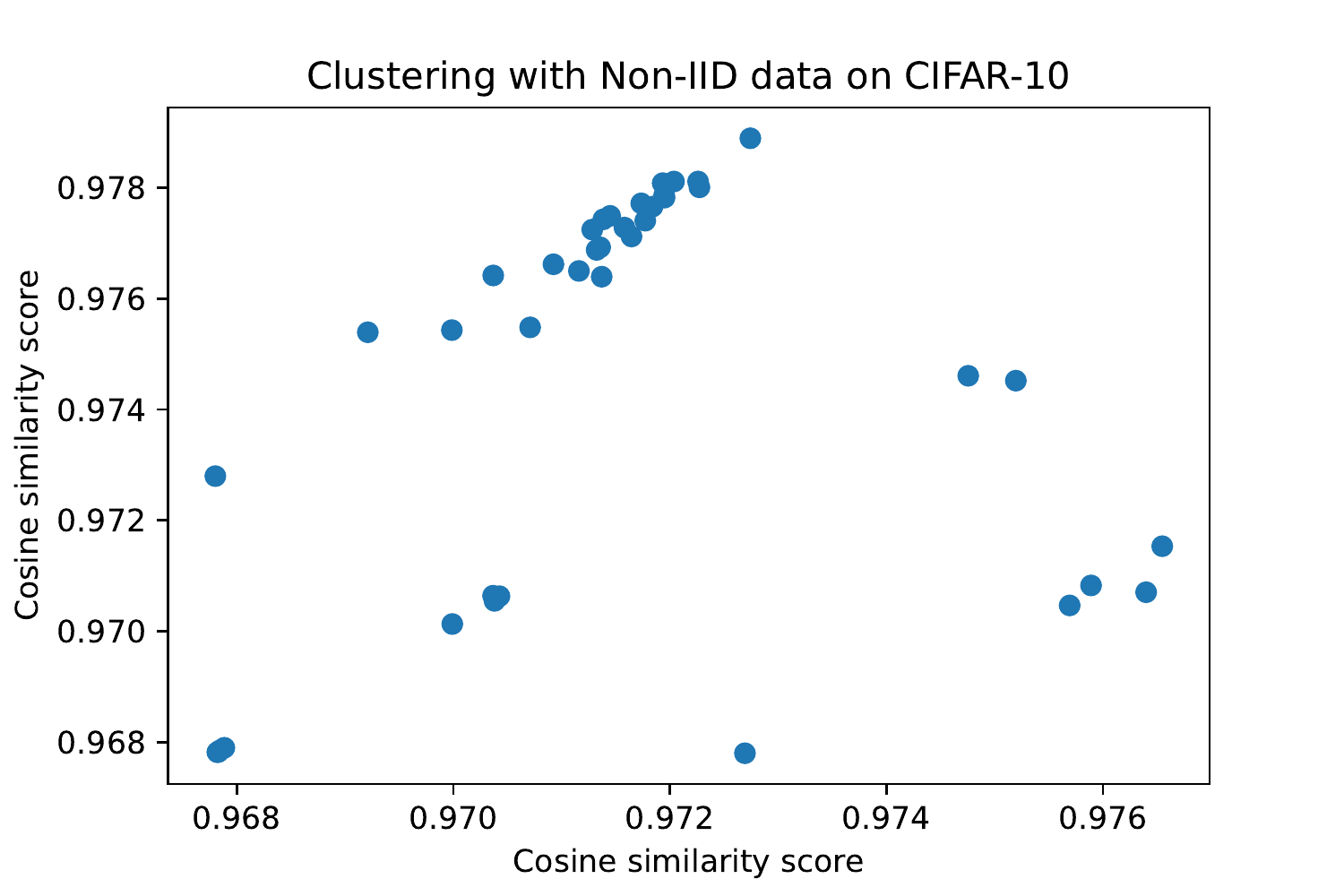}
  \end{minipage}
  \hfill
   \begin{minipage}[b]{0.24\textwidth}
    \includegraphics[width=\textwidth]{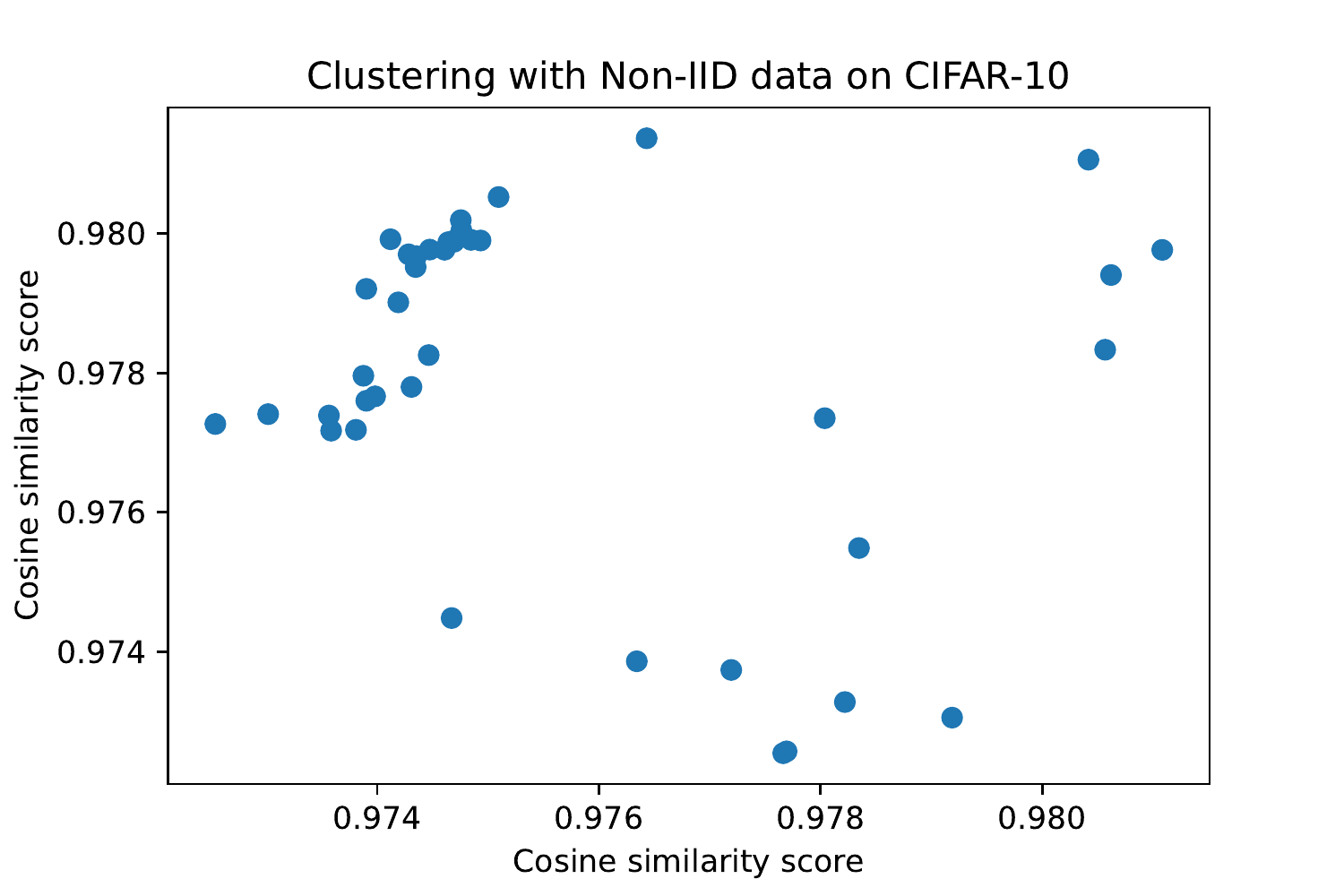}
  \end{minipage}
  \hfill
  \begin{minipage}[b]{0.24\textwidth}
    \includegraphics[width=\textwidth]{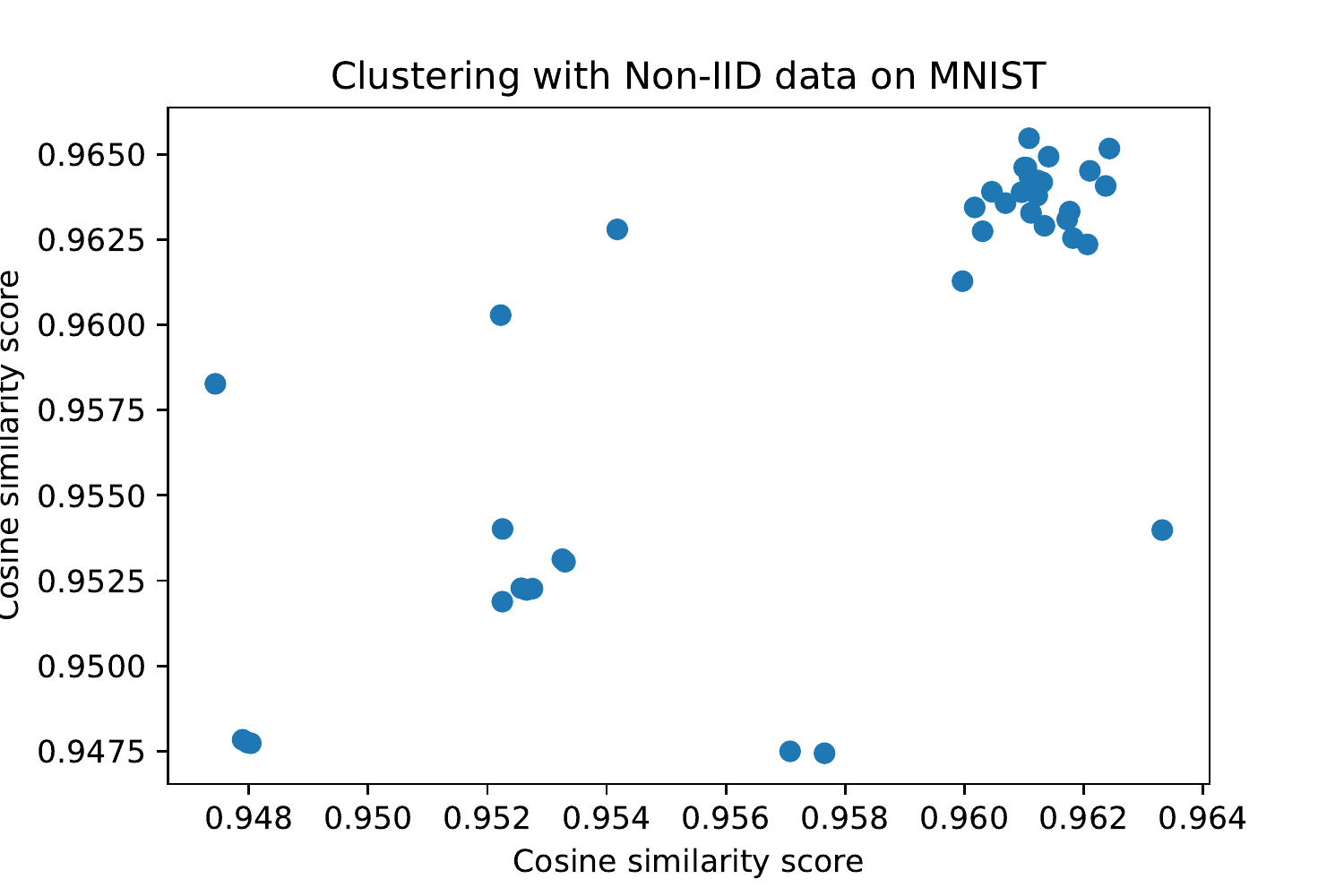}
  \end{minipage}
  \hfill
  \begin{minipage}[b]{0.24\textwidth}
    \includegraphics[width=\textwidth]{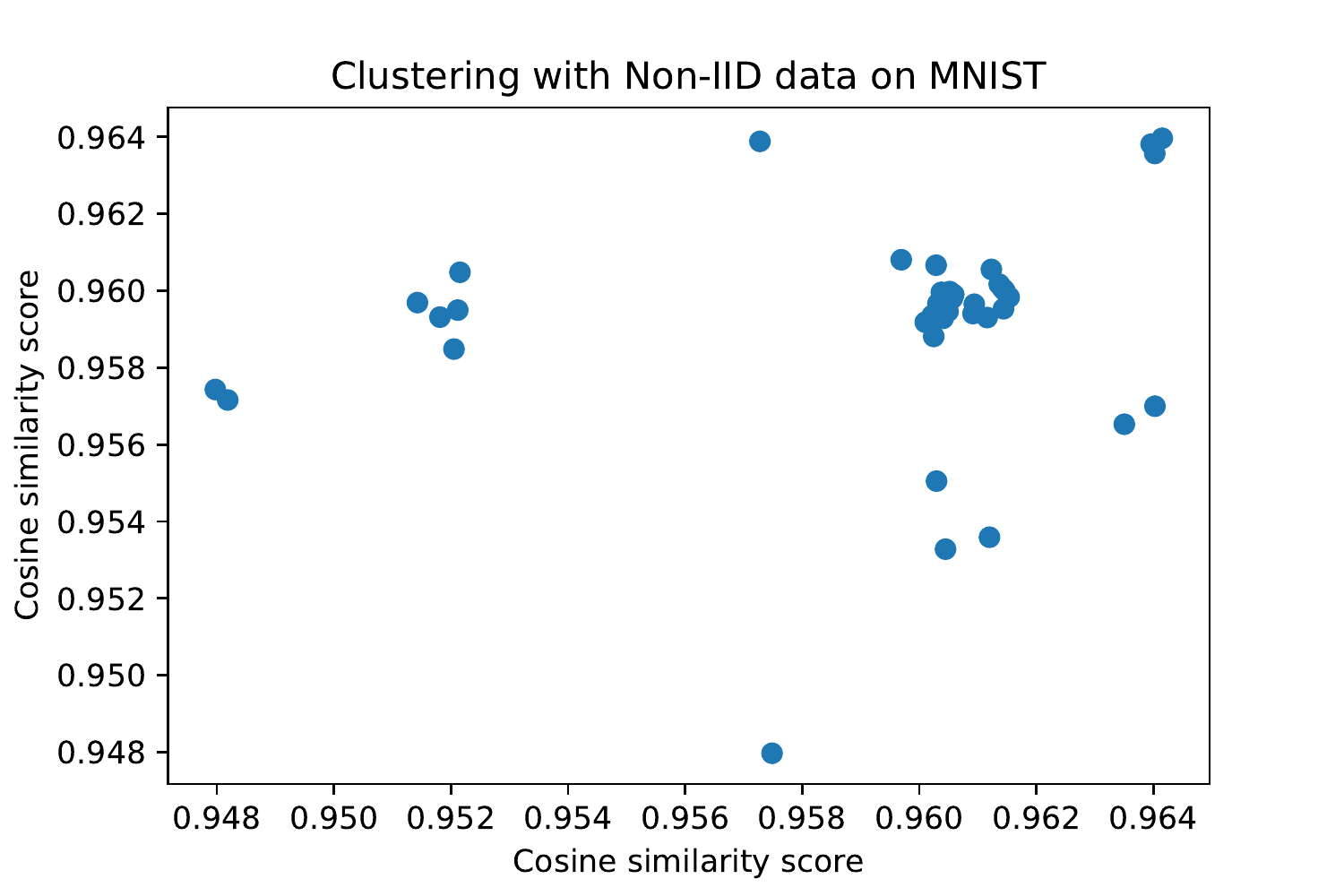}
  \end{minipage}
  \caption{FedLE in different levels of heterogeneity using CNN and MLP. First row: CNN model using CIFAR-10 with each client containing 2 and 3 classes of data. Second row: CNN model using MNIST with each client containing 2 and 3 classes of data. Third row: MLP model using MNIST with each client containing 2 and 3 classes of data.} \label{fig:cluster-hetero}
\end{figure}

We design a simple experiment to test the robustness of the similarity matrix using data with different levels of heterogeneity and imbalanced classes. For the clarity of demonstration, we design two levels of heterogeneity. For MNIST with CNN, MNIST with Multi-layer Perceptron (MLP), and CIFAR-10 with the CNN models, we used 40 clients and assign each client only 2 and 3 classes with 50\% clients sharing data from the same class. The weight-space graph in Fig. \ref{fig:cluster-hetero} shows that regardless of the level of data heterogeneity, their clustering pattern is clear.

Given the information, we are convinced that the building process of the similarity matrix can be a one-time procedure. 
Therefore, the communication cost from the clients to the server in building the similarity matrix is substantially reduced by the limited number of creating and updating the matrix and by the reduced size of model parameters. 
With each round of training, client selection is based on the matrix and its clusters. 
We can select clients with more diverse data in a single round which contribute more to the global model. 

Depending on how frequently and how much the data on each client changes, reconstruction of the matrix is likely necessary. 
If the client data is dynamic and constantly changing then there is a need for updating the weight matrix.
However, this is not the focal point of this paper.

\subsection{Forming the clusters}
The clustering operation takes place only once during the entire training process in order to maximize energy efficiency. All clients will perform one epoch of local training prior to uploading their local model updates. Each client's weight is compared with the weights of all other clients based on the local model weights. After the similarity matrix is built, FedLE picks the lowest $N_s$ similarity value from the matrix. The motivation behind picking the clients with the lowest similarity for clustering is to locate the clients with higher diversity and to avoid repeatedly selecting clients with lower similarity. In our experiments, we pick the lowest similarity value for analysis, which links to a pair of clients $c_\alpha$ and $c_\beta$. These two clients are considered the farthest from each other. Every other client  would have a similarity score in between. To tackle the overfitting problem on the majority of clients with similar data, we set different probabilities for every cluster based on their size. For the cluster with the most clients, we categorize that cluster as the majority cluster. The majority cluster has a much lower chance of being selected. The rest of the clusters have an equal probability of being selected. This way the clusters that carry statistically similar data would have a lower chance of being selected. Clusters with fewer clients will have a higher chance of being selected for training.

\subsection{Proposed FedLE Client Selection Method}
\label{sec:proposed-cs}
In this section, we propose a novel client selection method based on gradient similarity for FL to optimize training in an energy restraint scenario. We introduce two baseline methods and one optimized method. Based on our discussions in Section \ref{sec:similarity}, in FedLE, we only partially load model parameters to the server when updating the matrix. There is a significant difference between the local client models in terms of model weights and bias, particularly in the first and last layers of the CNN model. Instead of using the entire model to formulate the matrix, we only use the model parameters of the first convolution layer and the last fully connected layer.

\begin{algorithm}[htb]
\caption{The FedLE algorithm. \textit{K} is the number of participating clients. C $\in$ (0,1] is a hyperparameter that indicates the fraction of clients participating in each round of FL training.}\label{alg:Matrix client seleciton}
\begin{algorithmic}[1]
\State\textbf{Initialization:} Server sends global model to all clients.
\State\textbf{Constructing Similarity Matrix:} See Algorithm \ref{alg:Matrix}.
\State\textbf{K-means Clustering:} Server uses K-means clustering to cluster every client based on their pairwise similarity.
\State\textbf{Client Selection:} Server selects \textit{K$\times$C} clients based on the  cluster size.
\State\textbf{Distribution:} Server distributes the newest global model to selected clients this round.
\State\textbf{Update and Upload:} Selected clients update their local model parameter after local training using the global model and local data.
\State\textbf{Aggregation:} Server averages uploaded parameters and aggregates into the new global model for the next round.
\State All steps from \textbf{Client Selection} repeat until the model achieves the desired result.
\end{algorithmic}
\end{algorithm}

\section{SIMULATION SETUP}
\label{sec:setup}
In this section, we evaluate FedLE in a simulated energy-restraint environment by comparing FedLE with two baseline methods FedBO and FedAvg-B, using MNIST and CIFAR-10.

\subsection{Baseline Methods}
The first baseline approach is FedAvg with the consideration of battery constraints, named \textit{FedAvg-B}. In this baseline method, client selection follows a uniformly random method, the same as FedAvg, but only adding the battery level constraints to all clients. Clients with a battery level below the critical power level $\delta$ will be dropped during training.

The second baseline approach is named \textit{Battery-Only (FedBO)}. In FedBO, the client selection prioritizes clients with high battery levels. FedBO calculates each client's maximum training capacity based on their initial battery level. The probability of selecting each client is then generated based on its maximum capability. Essentially, low-powered clients have a lower chance of being selected than high-powered clients.

For our Battery-Optimized approach, prior to training, all available clients do local training for 1 epoch using their local data and upload part of the local model parameters to the server. The central server computes the similarity matrix and uses K-mean clustering. The new matrix will be used for client selection. We select clients based on the following criteria:
\begin{itemize}
    \item FedLE gives a higher probability to clusters with fewer clients and a lower probability to those with more clients.
    \item In each cluster, clients with higher battery levels have higher probabilities of being selected to optimize energy 
\end{itemize}

\subsection{Simulated Environment}
Consider a typical FL scenario with $K$ clients and a central server. Each client trains its local data based on the global model maintained by the server. 
We simulate an energy-constrained FL environment with imbalanced classes using the following variables. The training and communication costs are reflected in the remaining battery levels of the clients. The battery level \textit{b} of each client determines when the client would be excluded from FL training. Battery discharge \textit{r} applies to all clients after each round even if the client is not selected for training. Any other communication from clients to the server has additional communication cost \textit{a}.

Every client is initialized with battery level \textit{B}. Low-power clients have $b \in [0.245, 0.265]$, and high-power clients have $b \in [0.895, 0.905]$. We initialize 50\% of clients as low-powered clients and the other 50\% of clients as high-powered clients. When a client's battery level $b\leq \delta$, where $\delta = 0.2$, it will be considered to have reached the critical level of the battery and will be excluded from any further training.  All clients have a battery standby cost $r \in [0.1, 0.15]$. We also set the additional battery costs $s \in [0.4, 0.5]$ and the additional communication costs $a \in [0.24, 0.25]$. In the simulated environment, we assume that all clients have the same amount of data and computational power. There are 50\% clients assigned data points from the same class, creating an imbalanced class scenario. We also establish an early-stopping point for the experiment. When 50\% of the clients have their battery level below $\delta$, the entire FL training process stops.

\subsection{Non-IID \& imbalanced data and Global models}
 We adopt both CNN and MLP models in the evaluation. The same setting is used in the FL environment for all three methods for comparison. The FL settings for MNIST and CIFAR-10 are listed below and are according to the definitions in \cite{mcmahan2017communication}. 
 We use $K = 40$ and $C = 0.05$, which comes to exactly 2 clients per round of training. To create a simulated environment for non-IID FL and imbalanced classes, we divide the MNIST training set into 20 shards of size 3000 and the CIFAR-10 training set into 80 shards of size 625. Then we assign the first 20 clients with one shard each and the other 20 clients to the same class.




\section{Performance Evaluation} \label{sec:experiment}
In this section, we evaluate the performance of FedLE with two other baseline methods from Section \ref{sec:setup}.

\subsection{Test Accuracy}
\begin{figure}[t]
    \centering
    
    \vspace{-0.1in}
    \includegraphics[width=0.68\linewidth]{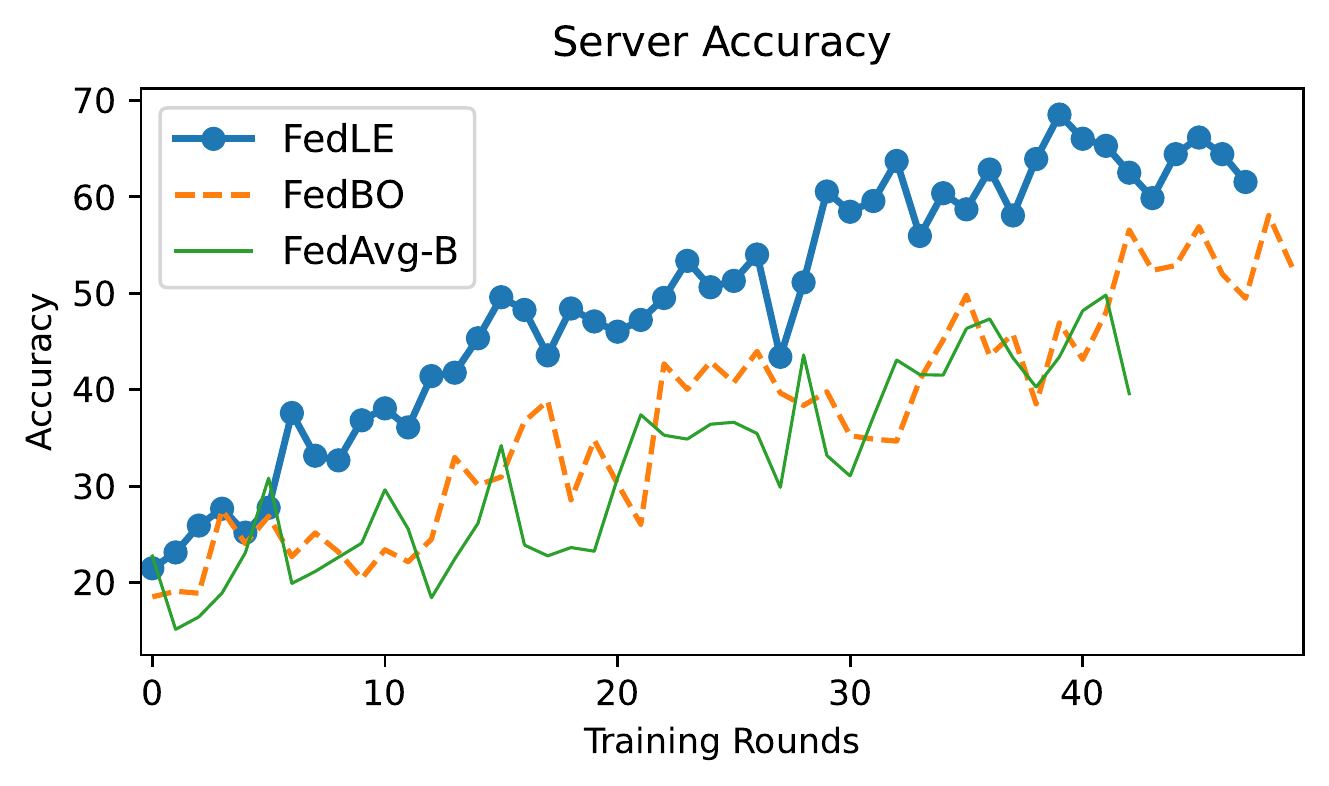}
    \includegraphics[width=0.68\linewidth]{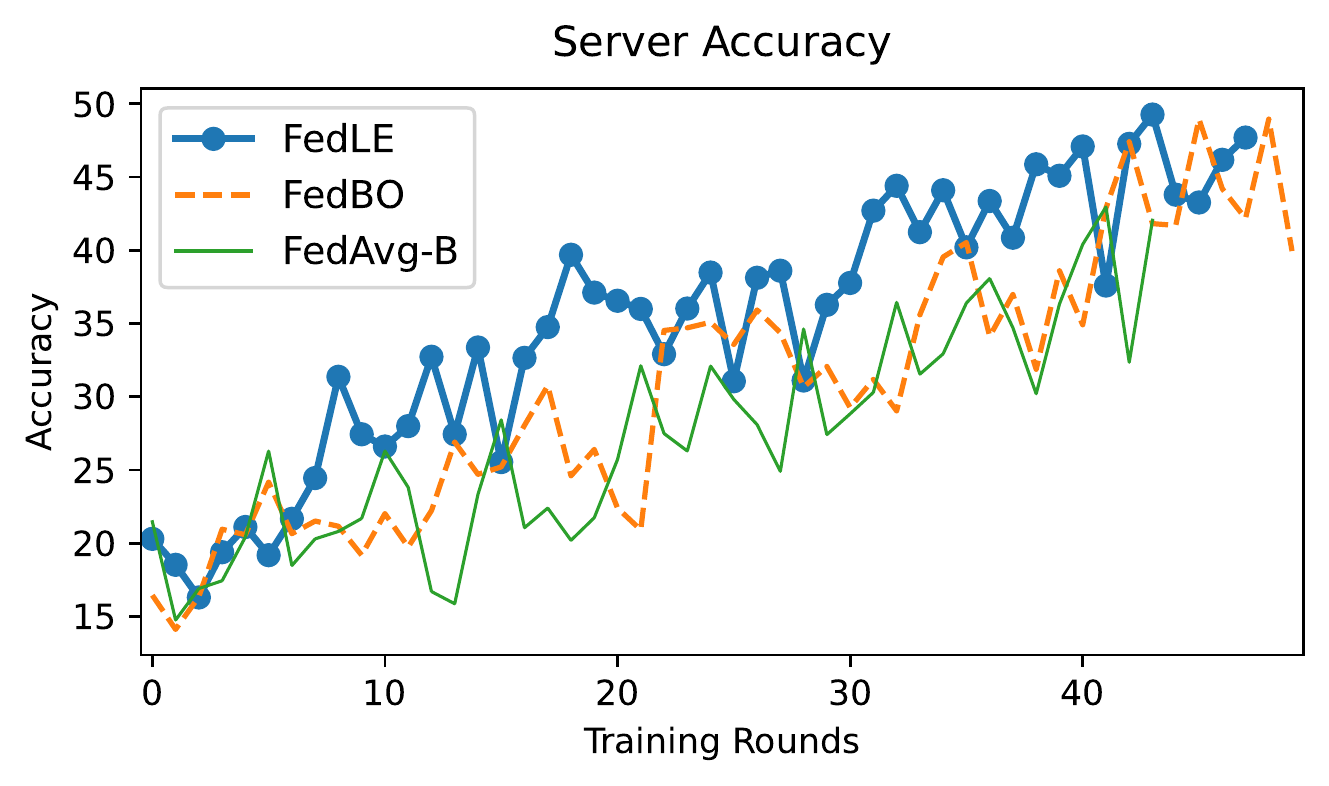}    
    \vspace{-0.1in}
    \caption{CNN and MLP on MNIST with FedLE, FedBO, and FedAvg-B.}
    \vspace{-0.1in}
    \label{fig:cnn-mnist}
\end{figure}

As shown in Fig. \ref{fig:cnn-mnist}, FedLE with a CNN model outperforms the other two baselines in test accuracy by a wide margin (around 10\%). 
A side-by-side comparison shows that FedLE works well when clients have non-IID data and imbalanced classes. 
The result shows that optimizing the client selection alone can have a significant effect on accuracy. 
We can see that FedLE is able to extend the training compared to FedAvg. 
We also tested our FedLE algorithm with another model. 
Using an MLP model, FedLE converges faster in the earlier rounds but only slightly outperforms the other two baselines toward the end of the training.
However, we believe it is due to the simplicity of the MLP model. 
Under the extreme Non-IID and imbalanced class scenario, the simple MLP model is difficult to converge, while the CNN model is much more resilient.

\begin{figure}[t]
    \centering
    \includegraphics[width=0.68\linewidth]{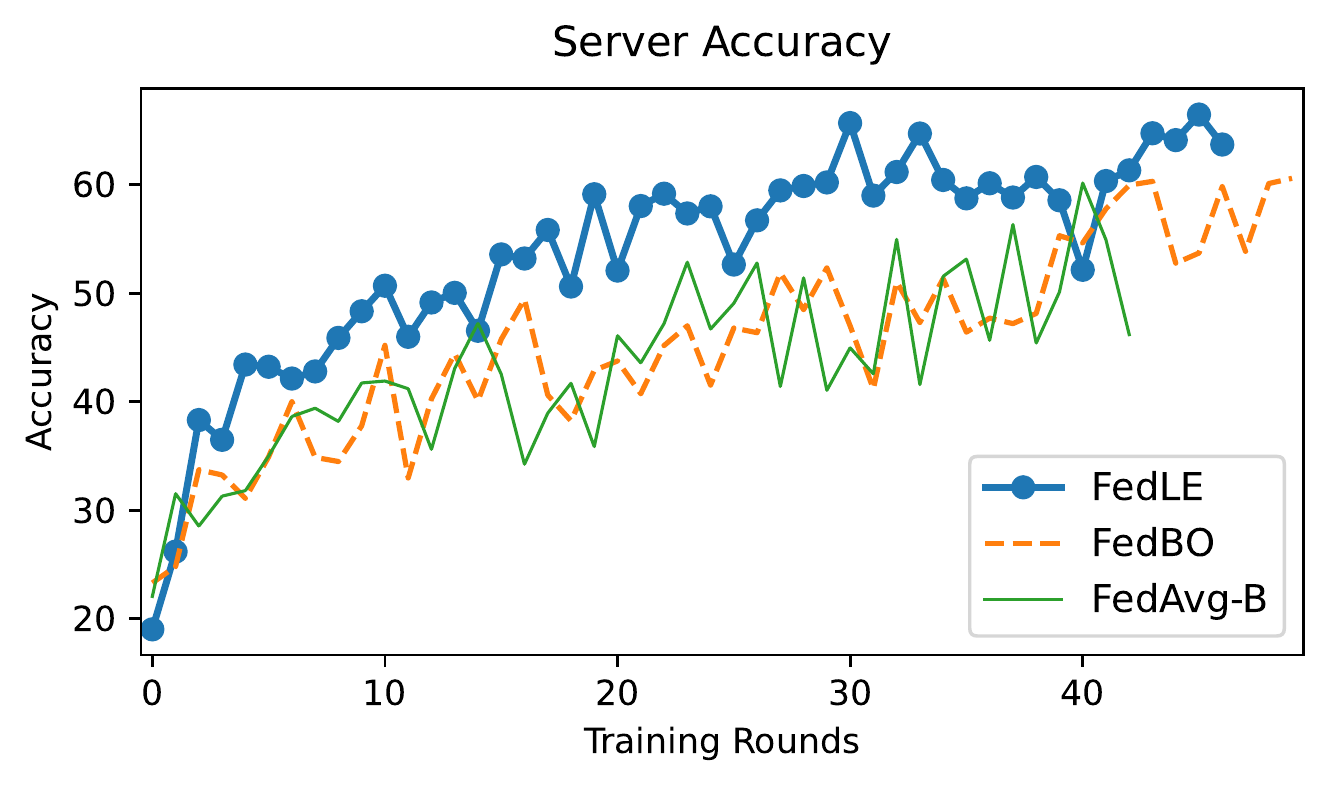}
    \vspace{-0.1in}
    \caption{CNN on CIFAR-10 with FedLE, FedBO, and FedAvg-B.}
    \label{fig:CIFAR-10 CNN}
    \vspace{-0.2in}
\end{figure}


\subsection{Number of Clusters And Majority Class}

\begin{figure}[htb]
    \centering
    \includegraphics[width=0.493\linewidth]{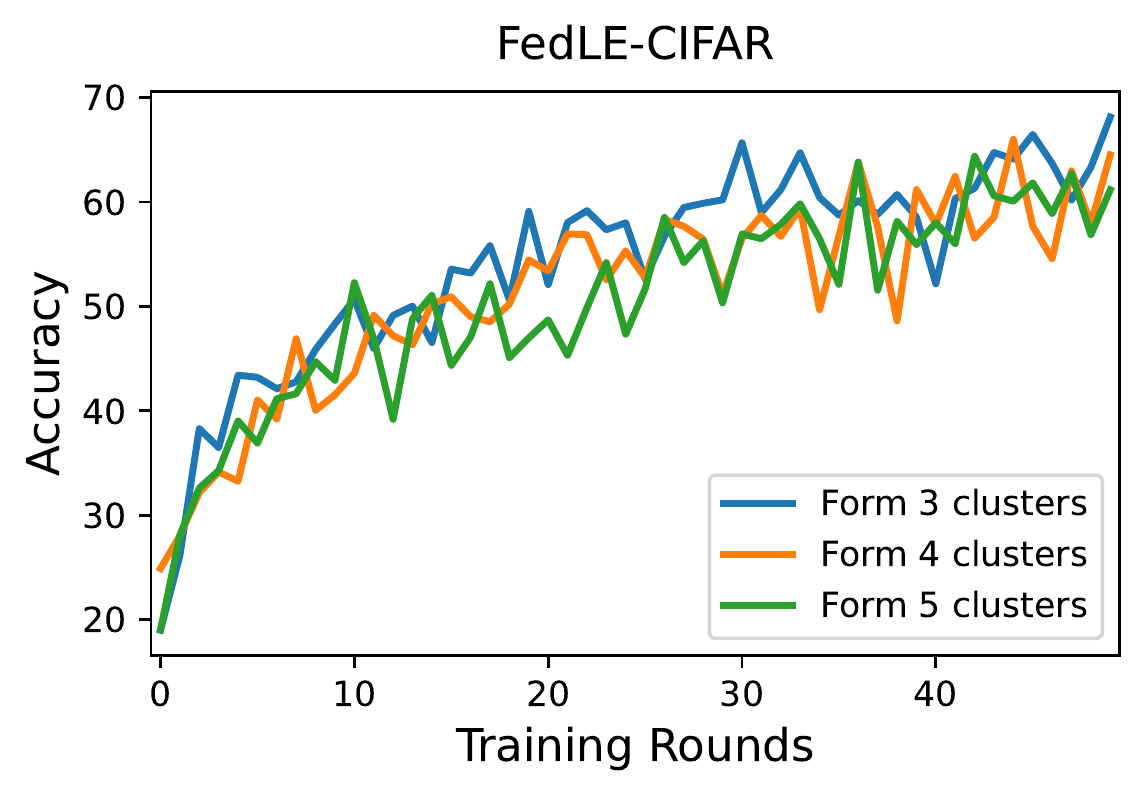}
    \includegraphics[width=0.493\linewidth]{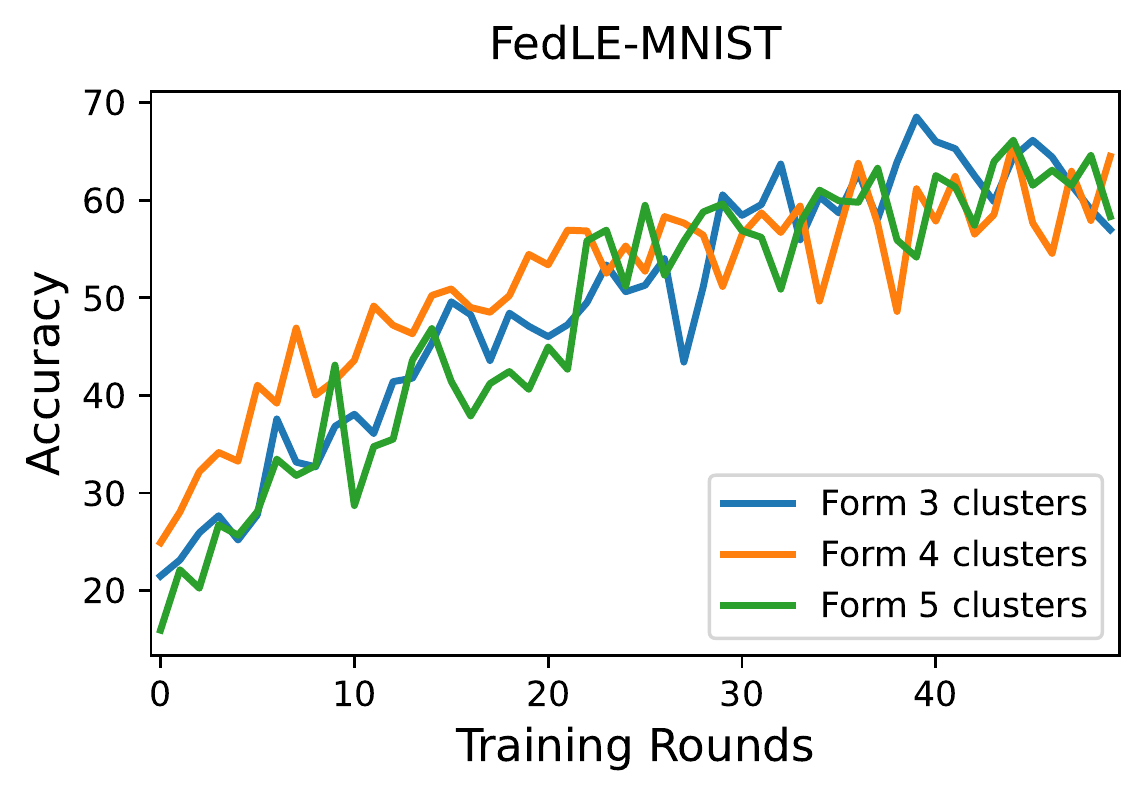} 
    \caption{FedLE using different cluster size for MNIST and CIFAR-10.}
    \label{fig:MNIST-cluster-size}
    \vspace{-0.1in}
\end{figure}

We run another experiment to test how the number of clusters affects the performance of FedLE.
As shown in Fig. \ref{fig:MNIST-cluster-size}, increasing the number of clusters does not have a significant effect on performance. 
We believe FedLE clusters majority class data together regardless of the number of clusters.
By increasing the number of clusters, we would only further divide the clusters' minority class data. 
Therefore, the impact on accuracy is low when increasing the number of clusters. 
From the results, we can see that the optimal number of clusters should be the sum of the number of the majority class data and the number of training clients per round.

The experiment runs using the same setting in Section \ref{sec:setup} (averaged results with 5 different seeds). Our proposed method with varying numbers of clusters in the MNIST experiment shows that when the cluster size is over the number of training clients (2 in this case), the results are close regardless of the number of clusters since the majority of class data are correctly categorized by our method. FedLE with different numbers of clusters on CIFAR-10 shows similar trends.

\renewcommand{\arraystretch}{0.8}
\begin{table}[t]

   \caption{Rounds lasted before reaching critical battery level $\delta$} 
   \label{table:rounds}
   \small
   \centering
   \begin{tabular}{lccc}
   \toprule
   {    } & {\textbf{FedLE}} & {\textbf{FedAvg-B}} & {\textbf{FedBO}} \\ 
   \midrule
   \emph{70\% Low-Powered Clients} & &&\\
   \midrule
   MNIST CNN  & 49 & 45 & 50 \\
   MNIST MLP & 48 & 44 & 50 \\
   CIFAR-10 CNN & 48 & 45 & 50 \\
   \midrule
   \emph{50\% Low-Powered Clients} & &&\\
   \midrule
   MNIST CNN  & 48 & 43 & 50 \\
   MNIST MLP & 49 & 44 & 50 \\
   CIFAR-10 CNN & 48 & 43 & 50 \\
   \midrule
   \emph{30\% Low-Powered Clients} & &&\\
   \midrule
   MNIST CNN  & 49 & 45 & 50 \\
   MNIST MLP & 48 & 44 & 50 \\
   CIFAR-10 CNN & 49 & 45 & 50 \\
   \bottomrule
   \end{tabular}
   \vspace{-0.15in}
\end{table}

\subsection{Energy saving results}

In Table \ref{table:rounds}, we show the performance of energy preservation in terms of the total number of rounds lasted for the edge IoT network. When 50\% of the clients become unavailable due to low power, we consider the network to be unusable, and the FL training stops. We can see while FedLE does not outperform FedBO which prioritizes solely on energy, it inches close with only 2 rounds short. Compared with the classic FedAvg-B, FedLE is able to extend the training for about 5 more rounds, or 11.6\%. Clustering the clients that share similar data distributions, we protect the low-powered clients by lowering the probability of selecting them and replacing them with high-powered clients within the same cluster. As there are still small chances low-powered clients are selected, we believe that is the reason why FedBO performs slightly better since it prioritizes the high-powered clients much more.

\section{Conclusions}
\label{sec:conclusion}
In this paper, we proposed FedLE, a clustering method based on the similarity matrix of client models to balance the energy consumption of battery-powered edge IoT clients. Experimental results verified that FedLE outperforms baseline methods while extending the lifespan of edge IoT networks by protecting low-powered clients from draining batteries. Future work includes studying the effect of other deep learning models, since the matrix is calculated based on model parameters.

\vspace{-0.05in}
\section*{Acknowledgement}
\noindent This material is based in part upon work supported by University of Calgary Start-up 10032260, National Science Foundation under Grant IIS-2212174, IIS-1749940, Office of Naval Research N00014-20-1-2382, and National Institute on Aging (NIA) RF1AG072449.

\bibliographystyle{IEEEtranN}
\bibliography{references}

\end{document}